%% file: main.tex
\documentclass[]{spie}  

\usepackage{comment}
\usepackage{booktabs}
\usepackage{amsmath,amsfonts,amssymb}
\usepackage{graphicx}
\usepackage{gensymb}
\usepackage{caption}
\usepackage{subcaption}
\usepackage{enumitem}
\usepackage[colorlinks=true, allcolors=blue]{hyperref}
\usepackage{listings}
\title{RGB-to-IR image translation for infrared vehicle detection in unseen UAV domains}
\author[a]{Thijs A. Eker}
\author[a]{Ella P. Fokkinga}
\author[a]{Jan Erik van Woerden}
\author[a]{Elfi I.S. Hofmeijer}
\author[a]{Sebastiaan P. Snel}
\author[a]{Klamer Schutte}
\author[a]{Friso G. Heslinga}
\affil[a]{TNO - Intelligent Imaging, Oude Waalsdorperweg 63, the Hague, the Netherlands}
\authorinfo{Corresponding author: Friso G. Heslinga. E-mail: fgheslinga@gmail.com}

\begin{document} 
\maketitle

\begin{abstract}
Synthetic training data has become an important resource for developing image-based AI systems in domains where measured data is scarce. This is particularly true for thermal infrared (IR) imagery, where the limited availability and diversity of real-world datasets constrain the training of IR-informed foundation models and subsequent finetuning for domain-specific tasks such as vehicle detection in air-to-ground scenarios. In contrast, large amounts of UAV-recorded RGB imagery are readily available, motivating the use of RGB-to-IR image translation to generate additional infrared training data. However, there is no one-to-one mapping between RGB and infrared appearance, as infrared signatures depend on physical characteristics that are not directly observable in RGB imagery (e.g., historical and engine-related heat signatures). This makes it challenging to learn transferable RGB-to-IR mappings.

In this work, we investigate whether generative RGB-to-IR image translation can improve infrared vehicle detection on previously unseen UAV datasets. RGB-to-IR translators are trained on paired RGB-IR imagery from multiple source datasets and applied to held-out target datasets to generate synthetic infrared training data. For each target dataset, RGB images from the training split are used as input to the RGB-to-IR translation models, while the corresponding infrared evaluation split is used only to evaluate vehicle detection performance. The evaluated methods include supervised GANs, ControlNet-based diffusion models, and foundation-model image editing using LoRA. The resulting synthetic IR imagery is used to train RF-DETR vehicle detectors. Experiments are conducted on five aerial RGB-IR vehicle datasets, with Kust4K and VTUAV serving as unseen target domains. Results show that synthetic IR consistently improves detection performance over simpler RGB and grayscale baselines. The best-performing approach, Stable Diffusion 3.5 with ControlNet, improves mAP from 50.8 to 60.1 on Kust4K and from 25.6 to 38.4 on VTUAV, compared to detectors trained only on source-domain infrared data. Further gains are observed on VTUAV by increasing output diversity through multiple diffusion seeds (+1.1 mAP) and prompt variations (+3.3 mAP). Although a gap remains to the upper bound obtained with real target-domain IR data, the results demonstrate that generative RGB-to-IR translation is an effective strategy for augmenting scarce infrared training data and improving cross-domain aerial vehicle detection.


\end{abstract}

\keywords{RGB-to-IR translation; ControlNet; Guided diffusion; Generative AI; Data augmentation}

\input{sections/01_introduction}

\input{sections/02_relatedworks}

\input{sections/03_methods}
\input{sections/04_results}

\input{sections/05_discussion}
\bibliography{report} 
\bibliographystyle{spiebib} 

\end{document}

%% file: sections/01_introduction.tex
\section{Introduction}
\label{sec:intro} 

Object detectors achieve strong performance across applications such as traffic monitoring, surveillance, autonomous systems, and military reconnaissance. Although recent advances, including foundation models, have reduced the amount of task-specific training data required in some fields \cite{liu2024groundingdino}, sufficient annotated data remains important for specialized domains such as infrared (IR) imagery, which are less well represented in common vision foundation models \cite{medeiros2025visual,moshtaghi2025rgbthbench}. Unlike RGB cameras, infrared sensors capture emitted thermal radiation. This allows objects to remain observable under challenging illumination conditions, including nighttime operation, adverse weather, and camouflage scenarios. These characteristics make infrared imagery particularly valuable for defense and security applications, where robust vehicle detection is required across a wide range of operating conditions.

Despite its operational relevance, the development of high-performance infrared object detectors remains constrained by the limited availability of annotated infrared datasets. Synthetic data provides one approach to address data scarcity and has shown promising results for training vehicle detectors in RGB imagery \cite{FrankAlma2024,Eker2023}. However, generating physically realistic infrared imagery using simulation is challenging. In contrast, large amounts of real RGB imagery are readily available and already capture representative scene geometry and variability. Consequently, RGB-to-IR image translation offers a promising approach for leveraging RGB data to increase the amount of infrared training data available for detector development.

However, generating realistic infrared imagery from RGB observations is fundamentally challenging. The relationship between visible appearance and thermal appearance is neither direct nor deterministic. Infrared images exhibit physical characteristics that are largely absent from RGB imagery, including engine heat signatures, thermal reflections, material-dependent heat retention, and temperature variations caused by historical environmental conditions. Many of these factors cannot be inferred reliably from RGB information. Furthermore, most existing RGB-to-IR translation methods rely on paired RGB-infrared training data collected within a specific dataset or operational context. In practice, paired infrared observations are often unavailable for the target scenario where synthetic data is needed, and translators frequently struggle to generalize to different sensors, environments, or viewpoints. This creates a domain adaptation challenge: can an RGB-to-IR translation model trained on a limited number of source datasets produce useful infrared imagery for previously unseen domains?

\begin{figure}[!t]
    \centering
    \includegraphics[width=\linewidth]{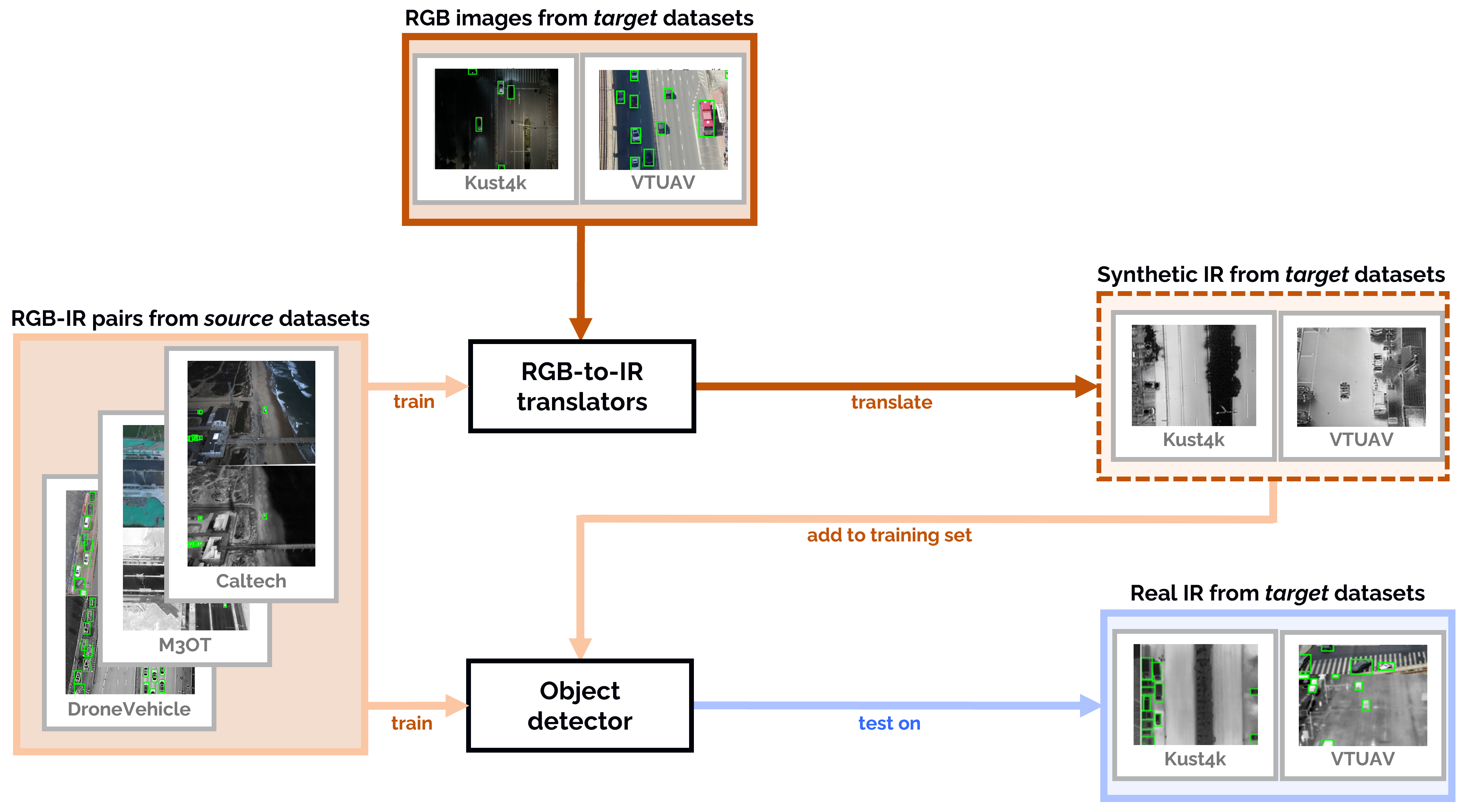}
    \caption{Overview of the proposed workflow. Paired RGB--infrared source datasets (DroneVehicle, Caltech, and M3OT) are used to train RGB-to-IR translation models. These models are subsequently applied to RGB imagery from the training splits of target datasets unseen during translation-model training (Kust4K and VTUAV) to generate synthetic infrared training data. The generated infrared imagery is added to the detector training set, and detection performance is evaluated on real infrared imagery from the separate target-domain evaluation splits. Orange indicates training data and blue indicates evaluation data.} \label{fig:overview_pipeline}
\end{figure}
In this work, we investigate whether generalized RGB-to-IR image translation can be used as a form of data augmentation for infrared vehicle detection. Figure~\ref{fig:overview_pipeline} provides an overview of the proposed training and evaluation procedure. Rather than training on a single dataset, we train RGB-to-IR translation models using paired data from multiple RGB-infrared datasets with the goal of learning infrared representations that transfer across domains. The resulting translation models are then used to generate synthetic infrared imagery from RGB images originating from previously unseen datasets. We evaluate whether adding these generated images to the detector training set improves vehicle detection performance in unseen infrared domains. In addition, we compare multiple RGB-to-IR translation approaches, ranging from conventional image-to-image translation methods to recent generative AI models, and contrast them with simpler baselines such as grayscale conversion. Through these experiments, we assess both the feasibility of cross-domain RGB-to-IR translation and its practical value for improving infrared air-to-ground vehicle detection.

Specifically, we (i) construct a multi-dataset RGB-infrared training dataset for air-to-ground vehicle detection, (ii) compare grayscale-based baseline with generative adversarial network (GAN) and diffusion-based methods for RGB-to-IR image translation, and (iii) evaluate the value of the generated training data for improving downstream vehicle detection performance on unseen infrared datasets.

%% file: sections/02_relatedworks.tex
\section{Related works}
\label{sec:relatedworks}

\subsection{Infrared UAV vehicle detection and domain shift}
Vehicle detection in UAV-based imagery is challenging due to small apparent size of objects, varying viewpoints and flight altitudes, complex backgrounds, and changing environmental conditions. Infrared imagery is frequently used in this context, because it is less dependent on visible illumination and can remain informative under low-light and nighttime conditions. Consequently, several UAV datasets containing infrared imagery and vehicle annotations have been introduced for the development and evaluation of infrared vehicle detection systems. These datasets cover a broad range of acquisition conditions and perception tasks, including vehicle detection, tracking, and semantic segmentation
\cite{sun2022dronevehicle,lee2025caltech,nie2025m3ot,ouyang2025kust4k,zhang2022vtuav,suo2023hituav,d2025saga}.

Collectively, these datasets span urban and natural environments, different times of day, flight altitudes, viewpoints, illumination conditions, sensors, and weather conditions. As a result, substantial domain shifts exist both within and between infrared UAV datasets. The Caltech Aerial RGB--thermal dataset highlights temporal and geographical domain shifts \cite{lee2025caltech}, while RTDOD formulates RGB--thermal UAV detection under incrementally changing domains, including changes in weather and illumination \cite{feng2023rtdod}. These studies highlight domain shift as a challenge in infrared UAV perception, making robust cross-domain generalization difficult \cite{Hofmeijer2026}.

\subsection{RGB-to-IR image translation}
Although several thermal UAV datasets have become available, collecting and annotating infrared imagery remains expensive and time-consuming. Because of this, infrared dataset availability is often insufficient for the specific domain in which a detector will ultimately be deployed. RGB-to-IR image translation provides a way to exploit more readily available RGB imagery and annotations for generating synthetic infrared training data. Early approaches mainly relied on GANs, with pix2pixHD providing a representative high-resolution conditional GAN approach \cite{wang2018pix2pixhd}. In the aerial domain, AVIID established a dedicated benchmark for visible-to-infrared translation \cite{han2023aviid}, while DR-AVIT introduced a GAN-based approach specifically targeting diverse and realistic aerial infrared generation \cite{han2024dravit}. Lee et al.\ further extended GAN-based translation with an edge-guided multi-domain approach that uses reference infrared images to control thermal appearance \cite{lee2023edgeguided}. 

More recently, diffusion-based methods have been applied to RGB-to-IR translation. \cite{fokkinga2025generative} ControlNet-based approaches adapt pretrained latent diffusion models using visible imagery as spatial conditioning \cite{bolanos2025eo2ir,reinhardt2025v2ir}. Other methods incorporate semantic and vision-language guidance to improve infrared translation \cite{ran2025diffv2ir,paranjape2026fvita}. Recent work has also adapted pretrained generative image models such as FLUX for cross-spectral translation using lightweight LoRA fine-tuning \cite{clouser2026fewshot}.

\subsection{Controllable and multi-domain infrared synthesis}
As discussed in the introduction, RGB-to-IR translation is a one-to-many problem, as thermal appearance is not uniquely determined by the corresponding RGB image. Diffusion-based image translation naturally supports this by producing different plausible outputs for the same conditioning image when different random seeds are used \cite{saharia2022palette}. Such stochastic variation can increase the diversity of synthetic training data, but provides no explicit control over which thermal appearance is generated. Recent approaches therefore introduce additional conditioning mechanisms to steer this variation toward particular infrared appearances.

Lee et al.\ encode a reference infrared image as a style representation, allowing the same RGB content to be translated with different thermal appearances \cite{lee2023edgeguided}, while ThermalGen explicitly disentangles thermal style to account for variations between sensors and acquisition conditions \cite{xiao2025thermalgen}. Text conditioning provides another mechanism for controlling the generated infrared domain. F-ViTA can generate long-wave IR (LWIR), mid-wave IR (MWIR), and near IR (NIR) translations from the same visible image \cite{paranjape2026fvita}, while TherA uses thermal-aware visual-language conditioning to control factors such as time of day, weather, and object thermal state \cite{lee2026thera}. In this work, we investigate whether simple dataset-specific captions can provide a lightweight way to learn and select dataset-associated thermal appearances from multiple training datasets (Section \ref{methods:rgb2ir}: dataset-specific \textit{prompt styles}).

\subsection{Synthetic infrared data for downstream perception}
The aforementioned methods focus on controlling the generated infrared appearance. However, the ultimate objective is often to improve downstream perception performance. Several works evaluate synthetic infrared imagery through its usefulness for downstream perception, including object detection, semantic segmentation, and tracking \cite{lee2023edgeguided,sikdar2024sslrgb2ir,zhang2019synthetic}. Lee et al.\ trained an object detector using translated infrared images and transferred RGB annotations, obtaining better detection performance than using alternative image translation methods \cite{lee2023edgeguided}. SSL-RGB2IR similarly demonstrates improvements in object detection and semantic segmentation when models are trained with synthetically generated infrared imagery \cite{sikdar2024sslrgb2ir}. ControlNet-based translation has also been used to generate synthetic infrared training data for object detection \cite{reinhardt2025v2ir}. In this work, Reinhardt et al.\ additionally adapt their translation model using infrared imagery from the target domain before generating synthetic training data. More recently, Clouser et al.\ adapt a flow-matching foundation model with LoRA and demonstrate improved infrared detection using translated RGB imagery \cite{clouser2026fewshot}, but also use paired target-domain RGB-infrared data to adapt the translator.

Although these studies show that translated infrared imagery can improve downstream perception, some approaches use target-domain infrared data during translation training or adaptation. We instead compare different RGB-to-IR translation approaches in unseen UAV domains where RGB imagery is available for synthetic data generation, while target-domain infrared imagery is entirely withheld from translator training.

%% file: sections/03_methods.tex
\section{Methods}
\label{sec:methods}
Figure~\ref{fig:overview_pipeline} provides an overview of our proposed workflow. Five paired RGB--infrared UAV datasets were collected and aligned (Section \ref{methods:datasets}). Then, RGB-to-IR translation models are trained on source-domain RGB--infrared pairs and applied to RGB imagery from unseen target domains to generate synthetic infrared training data (Section \ref{methods:rgb2ir}). Finally, the generated imagery was used to augment the training of an infrared vehicle detector, whose performance was evaluated on real infrared imagery from the target domains (Section \ref{methods:setup}). 

\subsection{Datasets}
\label{methods:datasets}
Five paired RGB-IR UAV datasets were used: DroneVehicle \cite{sun2022dronevehicle}, Caltech Aerial RGB-Thermal \cite{lee2025caltech}, M3OT \cite{nie2025m3ot}, Kust4K \cite{ouyang2025kust4k}, and VTUAV \cite{zhang2022vtuav}. Together, these datasets cover a wide range of sensors, image resolutions, flight heights, viewpoints, and scene types. All datasets were converted to a paired object-detection format containing aligned RGB images, infrared images, and vehicle bounding-box annotations. Since the downstream detection task was formulated as single-class vehicle detection, all retained vehicle categories were mapped to a single vehicle class. For all datasets, preprocessing was performed to obtain spatially aligned RGB-infrared pairs with consistent object-detection annotations. Following alignment, all annotations were manually verified. Missing bounding boxes were added, incorrect labels were removed, and regions that could not be labeled reliably were masked out. 

Table~\ref{tab:infrared_datasets} summarizes the infrared sensor characteristics and dataset statistics after preprocessing. The role of each dataset in training and evaluation is described in the experimental setup in Section~\ref{methods:setup}. Some characteristics were not reported in the original publications and are therefore marked as unknown. All datasets capture LWIR imagery in the 8--14~$\mu$m wavelength range. RGB-IR pairs were aligned with MatchAnything \cite{he2025matchanything}. Cross-modal correspondences were used to estimate an affine transformation with RANSAC, after which RGB images and RGB labels were warped into the infrared image coordinate system. Pairs were discarded when the alignment did not meet predefined quality criteria, including a minimum number of cross-modal correspondences and maximum allowed values for rotation, shear, and translation. These thresholds varied slightly between datasets. Rejected pairs most commonly occurred in low-light RGB imagery, where insufficient reliable correspondences could be found. A brief description of each dataset, together with any dataset-specific preprocessing steps, is provided below. 

\begin{table}
    \caption{Infrared camera characteristics and dataset statistics after preprocessing.
    \textit{Resolution} denotes the stored infrared image resolution; the native thermal
    sensor resolution is $640 \times 512$ for all datasets. \textit{Aligned pairs} denotes
    the number of RGB-infrared image pairs retained after preprocessing, alignment, and
    manual verification. The \textit{Used for} column indicates the role of each dataset
    in the experimental setup described in Section~\ref{sec:methods}: three source datasets
    (DroneVehicle, Caltech, M3OT) are used for RGB-to-IR translation and detector training,
    while two target datasets (Kust4K, VTUAV) are reserved for cross-domain evaluation.}
    \label{tab:infrared_datasets}

    \centering
    \scriptsize
    \setlength{\tabcolsep}{3pt}

    \begin{tabular}{
        @{}
        p{0.22\textwidth}
        p{0.19\textwidth}
        p{0.12\textwidth}
        p{0.10\textwidth}
        p{0.10\textwidth}
        p{0.20\textwidth}
        @{}
    }
        \toprule
        Dataset &
        Infrared camera &
        Flight height &
        Resolution &
        Aligned pairs &
        Used for \\
        \midrule

        DroneVehicle~\cite{sun2022dronevehicle} &
        \textit{Not specified} &
        80, 100, 120~m &
        $640 \times 512$ &
        23,529 &
        Training; in-domain test \\

        Caltech Aerial RGB-Thermal~\cite{lee2025caltech} &
        FLIR ADK &
        Mostly 40~m &
        $960 \times 600$ &
        223 &
        Training; in-domain test \\

        M3OT~\cite{nie2025m3ot} &
        Mavic 3T microbolometer &
        100--120~m &
        $640 \times 512$ &
        952 &
        Training; in-domain test \\

        Kust4K~\cite{ouyang2025kust4k} &
        Uncooled VOx sensor &
        \textit{Not specified} &
        $640 \times 512$ &
        2,279 &
        Close-domain test \\

        VTUAV~\cite{zhang2022vtuav} &
        Zenmuse H20T &
        5--20~m &
        $1920 \times 1080$ &
        1,577 &
        Far-domain test \\

        \bottomrule
    \end{tabular}
\end{table}
\subsubsection{DroneVehicle}
\begin{figure}[htbp]
\centering
\includegraphics[
width=\linewidth,
trim=0 1.5cm 0 4.1cm,
clip
]{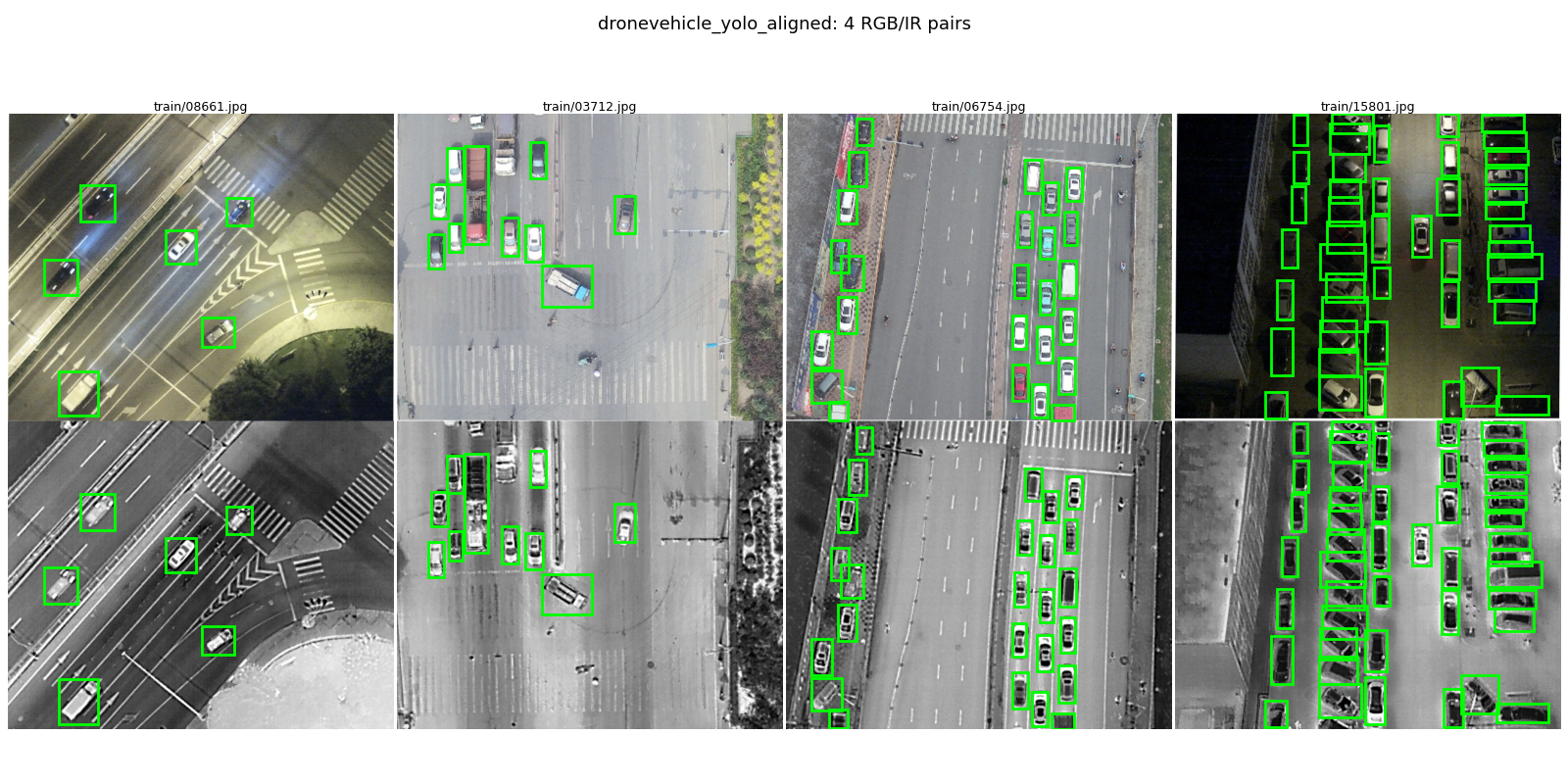}
\caption{Example aligned RGB-infrared pairs from DroneVehicle after cropping and alignment. Top row shows RGB images, bottom row shows IR images.}
\label{fig:dronevehicle_example_pairs}
\end{figure}

The DroneVehicle dataset~\cite{sun2022dronevehicle} is the largest dataset used in this study, containing UAV imagery at a resolution of 840 $\times$ 712 pixels. Before alignment, both modalities were center-cropped to $640\times512$ pixels to remove a white border surrounding the image content. After alignment, 23,529 paired samples were retained. The dataset contains many vehicles and mostly small bounding boxes, with most objects below 100 $\times$ 100 pixels. Example aligned DroneVehicle RGB-infrared pairs are shown in Figure~\ref{fig:dronevehicle_example_pairs}.

\subsubsection{Caltech}
\begin{figure}[htbp]
\centering
\includegraphics[
width=\linewidth,
trim=0 4cm 0 6.5cm,
clip
]{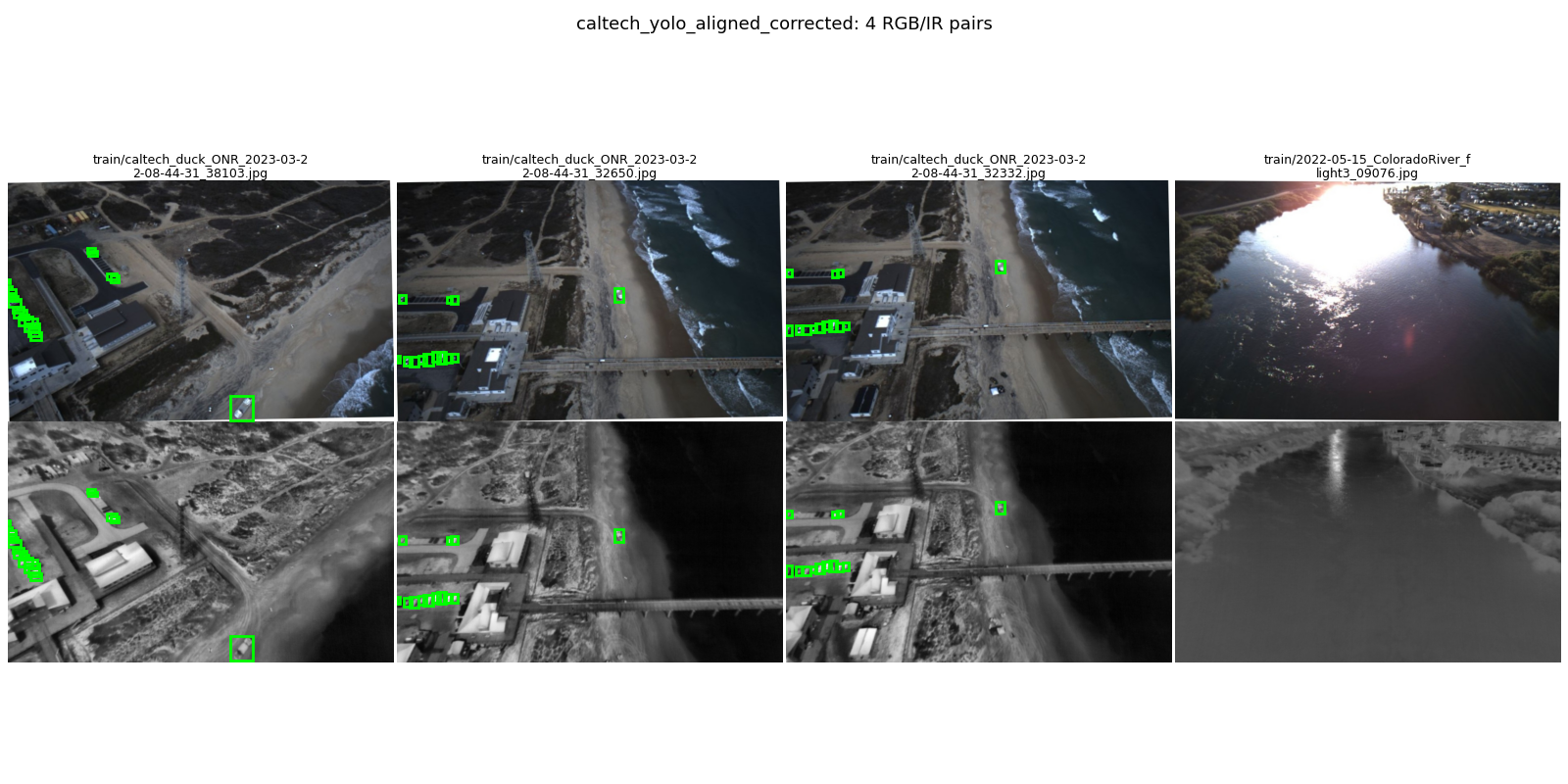}
\caption{Example aligned RGB-infrared pairs from the Caltech Aerial RGB-Thermal dataset after alignment and manual bounding-box correction.}
\label{fig:caltech_example_pairs}
\end{figure}

The Caltech Aerial RGB-Thermal dataset~\cite{lee2025caltech} contains aerial imagery at a resolution of 960 $\times$ 600 pixels. All frames containing annotated vehicles were selected together with an equal number of vehicle-free frames. Vehicle segmentation masks were converted to bounding-box annotations and manually verified after RGB-infrared alignment. This dataset is substantially smaller, with 223 paired samples remaining after matching and manual correction, but provided additional diversity through different backgrounds and environmental conditions. Example Caltech RGB-infrared pairs after alignment and bounding-box correction are shown in Figure~\ref{fig:caltech_example_pairs}.

\subsubsection{M3OT}
\begin{figure}[htbp]
\centering
\includegraphics[
width=\linewidth,
trim=0 1.5cm 0 4.1cm,
clip
]{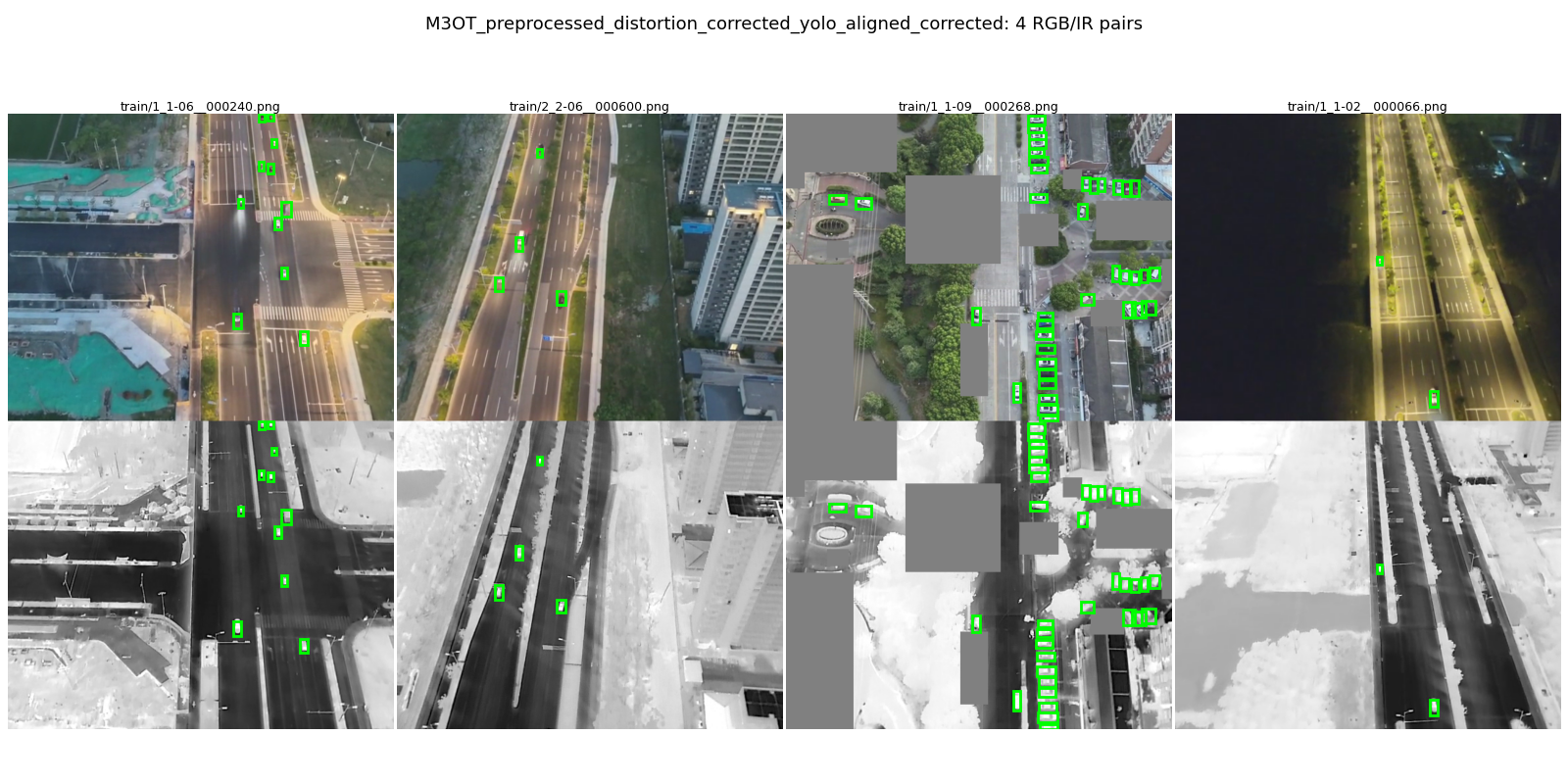}
\caption{Example aligned RGB-infrared pairs from M3OT after subsampling, radial distortion correction, alignment, and manual masking. Masked areas are indicated by the grey boxes.}
\label{fig:m3ot_example_pairs}
\end{figure}

M3OT contains paired UAV imagery captured at 640 $\times$ 512 pixels. A radial distortion correction was applied to the infrared imagery prior to alignment, and improperly annotated image regions were manually masked. Due to limited variation between consecutive frames, the dataset was subsampled at every tenth frame, resulting in 952 retained pairs. Similar to DroneVehicle, the vehicle bounding boxes are generally small due to the high flight altitude of $100-120m$. Example M3OT RGB-infrared pairs after subsampling, distortion correction, and manual masking are shown in Figure~\ref{fig:m3ot_example_pairs}.


\subsubsection{Kust4K}
\begin{figure}[htbp]
\centering
\includegraphics[
width=\linewidth,
trim=0 1.5cm 0 4.1cm,
clip
]{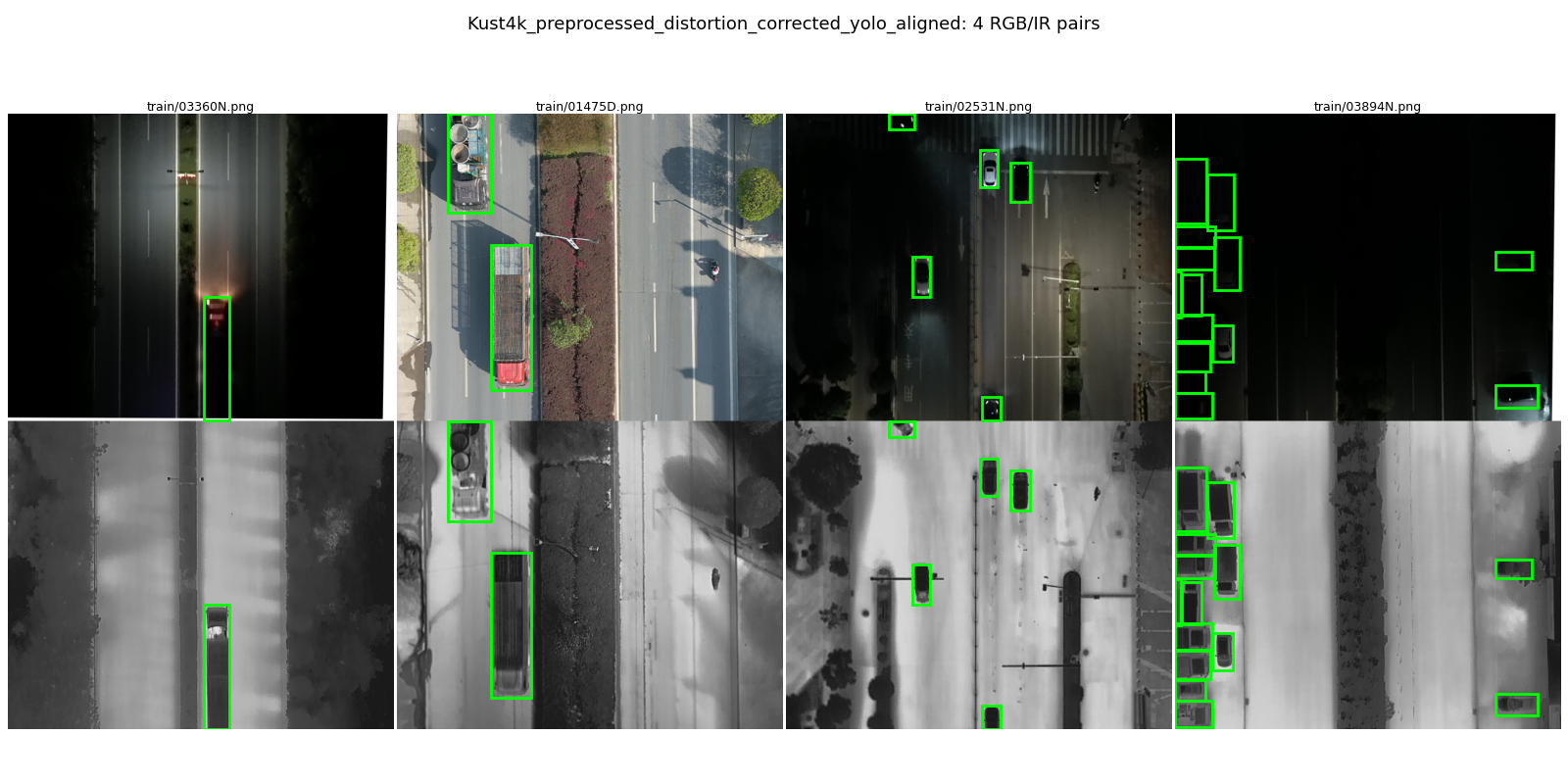}
\caption{Example aligned RGB-infrared pairs from Kust4K after radial distortion correction and alignment.}
\label{fig:kust4k_example_pairs}
\end{figure}

Kust4K contains UAV imagery at 640 $\times$ 512 pixels and is visually similar to DroneVehicle and M3OT in terms of viewpoint. Radial distortion correction was applied to the infrared images before alignment. After alignment and filtering, 2,279 pairs were retained. Example Kust4K RGB-infrared pairs after distortion correction and alignment are shown in Figure~\ref{fig:kust4k_example_pairs}.


\subsubsection{VTUAV}
\begin{figure}[htbp]
\centering
\includegraphics[
width=\linewidth,
trim=0 4.5cm 0 7.4cm,
clip
]{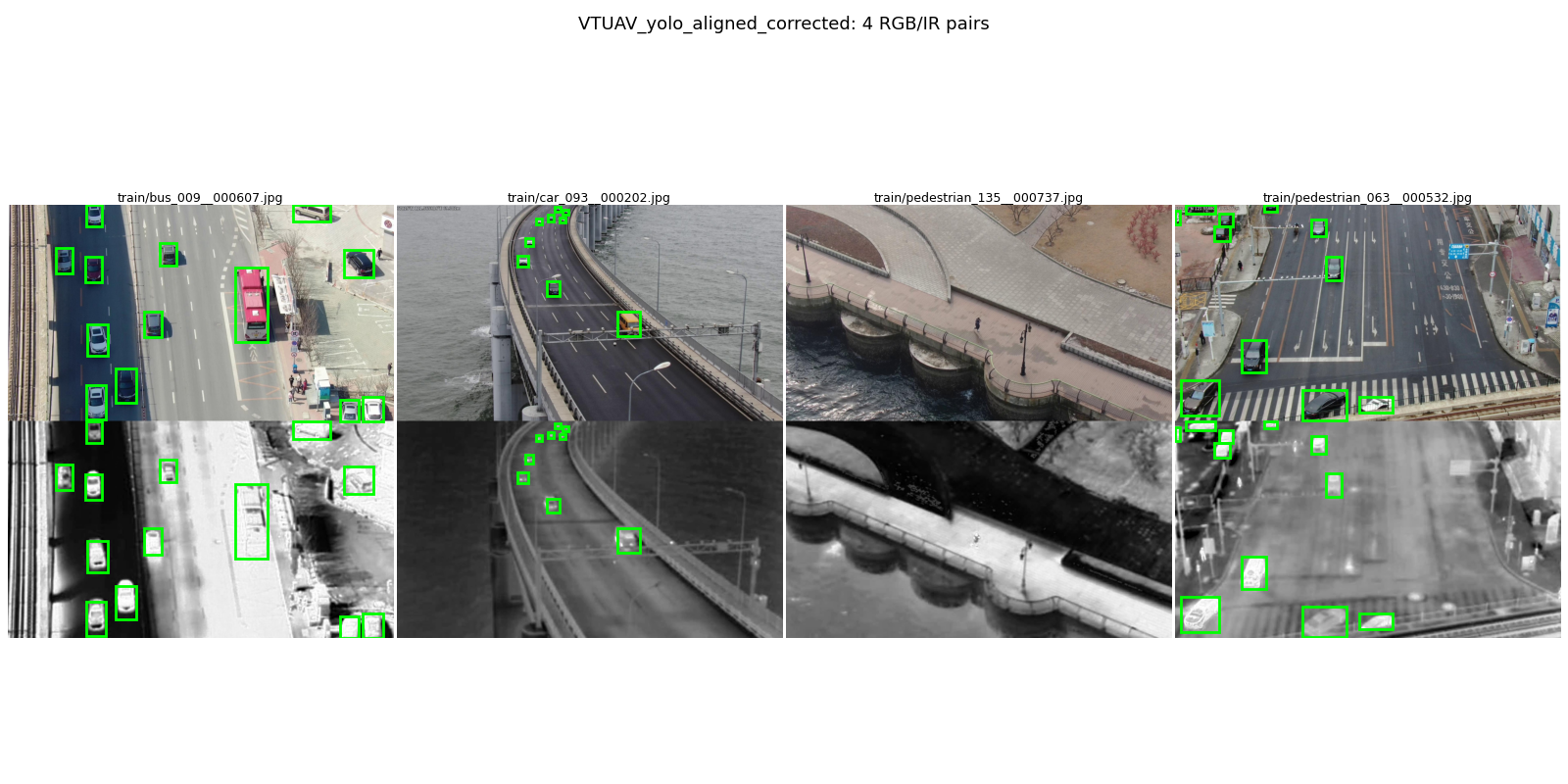}
\caption{Example aligned RGB-infrared pairs from VTUAV after sequence sampling, alignment, GroundingDINO pre-labeling, and manual bounding-box correction.}
\label{fig:vtuav_example_pairs}
\end{figure}

VTUAV differs more strongly from the source datasets (DroneVehicle, Caltech, and M3OT) than Kust4K. It contains UAV imagery at 1920 $\times$ 1080 pixels, collected from a lower flight altitude, resulting in larger apparent object sizes and a different viewing geometry. Furthermore, as VTUAV is a tracking dataset, the original annotations follow only a single object per sequence. To make the dataset suitable for object detection, a subset of frames is sampled from each sequence and automatically pre-labeled using GroundingDINO \cite{liu2024groundingdino}. The annotations are subsequently verified and corrected manually, after which the RGB-infrared pairs and labels are aligned using the same procedure described above. This resulted in 1,028 training pairs and 549 test pairs. Example VTUAV RGB-infrared pairs after sampling, alignment, prelabeling, and manual correction are shown in Figure~\ref{fig:vtuav_example_pairs}.


\subsection{RGB-to-IR translation methods}
\label{methods:rgb2ir}
Multiple RGB-to-IR translation approaches are evaluated, covering three major families of image-to-image generation methods: ControlNet-conditioned diffusion models, supervised GAN-based translation, and instruction-based foundation-model image editing. A simple grayscale conversion is included as a fourth, non-generative baseline. For methods requiring a different input size, images are resized with aspect-ratio preservation and padded to the required dimensions. After translation, the padding is removed and the output is resized back to the original image dimensions.

\paragraph{ControlNet-based translation.}
ControlNet~\cite{zhang2023adding} introduces additional spatial conditioning to a pretrained generative model, allowing an RGB image to guide the structure of the generated output. A similar ControlNet-based approach has previously been applied to RGB-to-IR translation by Bolanos et al.~\cite{bolanos2025eo2ir}. In our setup, the RGB image is used as the conditioning input and the target modality is specified using the fixed prompt \textit{``Aerial infrared photo.''}. We used two pretrained generative backbones in combination with ControlNet: Stable Diffusion 3.5 Medium~ (SD) \cite{esser2024scaling} and FLUX.1-dev~ (FLUX) \cite{flux2024}. For both backbones, only the ControlNet parameters are optimized while the pretrained base model, text encoders, and variational autoencoder remain frozen. The models are trained for 60,000 steps with an effective batch size of 8 and a learning rate of $1\times10^{-5}$.

Different random seeds can produce different infrared translations for the same RGB image. We hypothesize that using multiple independently generated variants can increase the diversity of the detector training data and improve downstream detection performance. For both SD and FLUX, we therefore compare training with one generated variant per RGB image against training with three variants generated using different random seeds.

\textit{Stable Diffusion 3.5 Medium.}
Stable Diffusion 3.5 Medium (SD)~\cite{esser2024scaling} is evaluated as the first ControlNet backbone. From this point onward, this method is referred to as \textit{SD-ControlNet}. At inference, SD uses 20 sampling steps, a conditioning scale of 1.0, and a guidance scale of 7.0.

We additionally investigated dataset-specific \textit{prompt styles} for SD. During training, each image pair is associated with a prompt of the form \textit{``Aerial infrared photo in the style of $<dataset>$''}, where the dataset identifier corresponds to the source dataset from which the pair originates. This provides the model with information about dataset-specific imaging characteristics and allows it to learn dataset-dependent RGB-to-IR mappings. At inference, each target-domain RGB image is translated using the prompt styles of all three source datasets, producing three infrared variants. We hypothesized that this increases appearance diversity and improves detector robustness by exposing the detector to multiple plausible infrared interpretations of the same scene.

\textit{FLUX.1-dev.}
FLUX.1-dev (FLUX)~\cite{flux2024} is evaluated as the second ControlNet backbone. From this point onward, this method is referred to as \textit{FLUX-ControlNet}. Apart from the seed-variation experiment described above, no prompt-style conditioning is applied for this backbone. At inference, FLUX uses 28 sampling steps, a conditioning scale of 1.0, and a guidance scale of 3.5.

\paragraph{pix2pixHD.}
pix2pixHD~\cite{wang2018pix2pixhd} is a supervised paired image-to-image translation method. The model is trained to map aligned RGB images directly to their corresponding infrared images using the paired source-domain training data. The model is trained for 100 epochs with a batch size of 4 and a learning rate of $2\times10^{-4}$.

\paragraph{FLUX.2 Klein with LoRA image editing.}
Stein et al., show that a domain shift can also be introduced with an diffusion based editing model.\cite{Stein2026}. Therefore, FLUX.2 [klein] 9B Base~\cite{flux2kleinbase9b2026} was fine-tuned as a representative foundation-model adaptation approach for paired RGB-to-IR image editing using low-rank adaptation (LoRA)~\cite{hu2021lora}. From this point onward, this method is referred to as \textit{FLUX-LoRA}. The model is trained exclusively on paired RGB--IR images from the three source datasets using the fixed instruction \textit{``convert this RGB drone image to thermal infrared.''} This approach is related to Clouser et al.~\cite{clouser2026fewshot}, who adapt FLUX.1 Kontext with LoRA for few-shot cross-spectral RGB-to-IR translation and subsequently use the generated imagery for infrared object detection. In contrast, our adapter is trained on the full source-domain training set and is evaluated by translating RGB imagery from target domains whose infrared imagery is entirely withheld during translator training.

The configuration uses rank-16 LoRA layers with a LoRA alpha of 16 and bf16 precision. The base model and text encoder remain frozen, and only the LoRA parameters are optimized. Training is configured for 21,000 optimization steps with a batch size of 4, a learning rate of $1\times10^{-4}$, and AdamW 8-bit optimization using 512- and 768-pixel resolution buckets. At inference time, target-domain RGB training images are provided as input images together with the same instruction. Images are generated at $640\times512$ pixels using 30 inference steps, a guidance scale of 4, and a LoRA scale of 1.0.

\subsection{Experimental setup}
\label{methods:setup}
The five datasets serve different roles in the experimental setup. DroneVehicle, M3OT, and Caltech are used as source datasets, providing paired RGB-infrared data for training the RGB-to-IR translation models and the baseline infrared object detector. Kust4K and VTUAV are held out during translator training and serve as target datasets. Kust4K is considered a \textit{close-domain} target dataset because of its similarity in viewpoint and image characteristics to the source datasets, whereas VTUAV represents a \textit{far-domain} target dataset, featuring imagery captured from substantially lower flight altitudes.

The baseline detector (Section \ref{methods:rfdetr}) is trained exclusively on infrared imagery from the source datasets and evaluated on three test scenarios: the in-domain source test sets, the held-out Kust4K test set, and the held-out VTUAV test set. This establishes performance when no target-domain data are used for detector training.

The RGB images from the target-domain datasets Kust4K and VTUAV train split are translated into synthetic infrared imagery and added to the source-domain infrared training data. For comparison, two simpler alternatives are considered that use the same target-domain RGB images directly, either as RGB images or converted to grayscale. The latter serves as a simple infrared proxy that preserves scene geometry without requiring a learned translation model. 

In addition, we evaluated whether grayscale and generative translation provide complementary benefits by training detectors on combinations of grayscale and SD-ControlNet-generated imagery. Three combinations are considered: single-seed SD-ControlNet with grayscale augmentation, multi-seed SD-ControlNet with grayscale augmentation, and prompt-style SD-ControlNet with grayscale augmentation. Finally, we performed an upper-bound experiment in which the real target-domain infrared imagery is added during training. 

\subsubsection{Object detector}
\label{methods:rfdetr}
RF-DETR is used as the downstream vehicle detector for all experiments, because it combines a modern transformer-based detector with a strong pretrained visual backbone, provides fast inference, and a straightforward fine-tuning workflow. RF-DETR is a real-time DETR-style transformer detector developed by Roboflow, using DINOv2 as its vision backbone~\cite{roboflow2026rfdetr}. All experiments are based upon RF-DETR Medium and the detector is trained for a single class: vehicle. Each detector configuration is trained for 20 epochs and uses 5 warm-up epochs. For experiments that increase the amount of training data through multiple generated variants per image, such as the multi-seed, prompt-style, or combined experiments, the number of training and warm-up epochs is reduced proportionally to ensure that the total number of training iterations remains approximately constant. The model is always initialized from pretrained RF-DETR Medium weights and the experiments are repeated using three random seeds. Images are resized to an input resolution of $640 \times 640$ when needed, and the batch size is 12. The main learning rate is $1\times10^{-4}$ and the encoder learning rate is $1.5\times10^{-4}$. The default augmentation setup for infrared drone imagery applies horizontal flipping, small brightness and contrast perturbations, and mild affine transformations. Additionally, random image inversion is included during training, so the detector sees both white-hot and black-hot style infrared appearances equally.

The evaluation is performed at the downstream task level using mean Average Precision (mAP) on real infrared test imagery from the target-domain datasets Kust4K and VTUAV. We report the COCO-style mAP$_{50:95}$ metric, averaged over IoU thresholds from 0.50 to 0.95 in steps of 0.05 \cite{lin2014microsoft}, and averaged over three detector training seeds.

%% file: sections/04_results.tex
\section{Results}
\begin{figure}
\centering
\includegraphics[width=\linewidth]{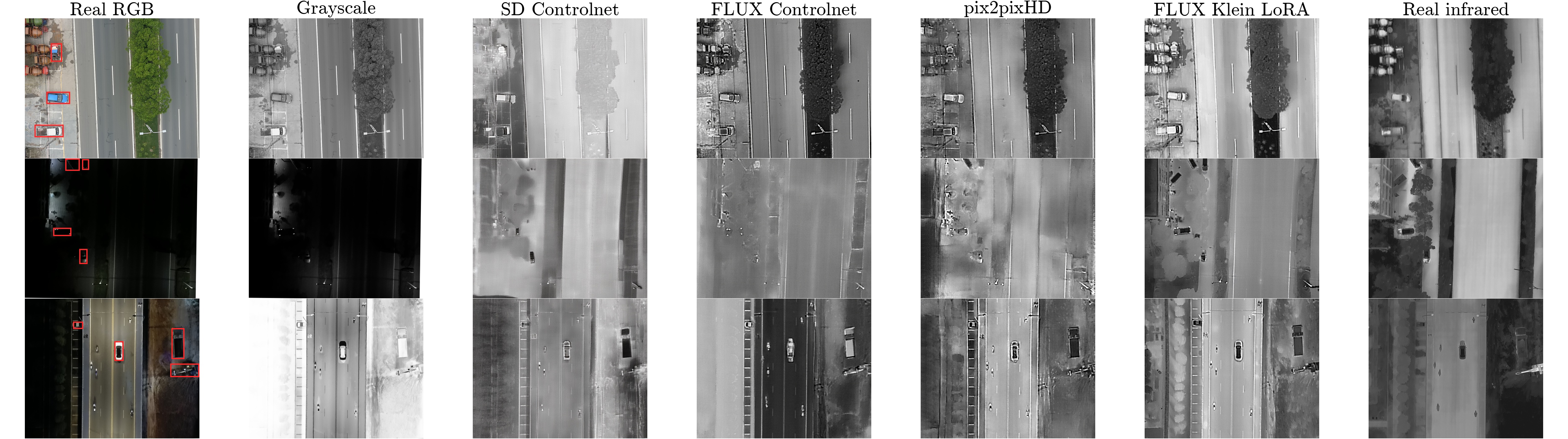}
\vspace{-3mm}
\caption{Three RGB-to-IR translation example scenes of the Kust4K training set. From left to right, each row shows an RGB image, the translations generated by the grayscale baseline, SD-ControlNet, FLUX-ControlNet, pix2pixHD, and FLUX-LoRA, and the corresponding real infrared images.}
\vspace{2mm}
\label{fig:kust4k_inference_examples}
\end{figure}

\begin{figure}
\centering

\includegraphics[width=\linewidth]{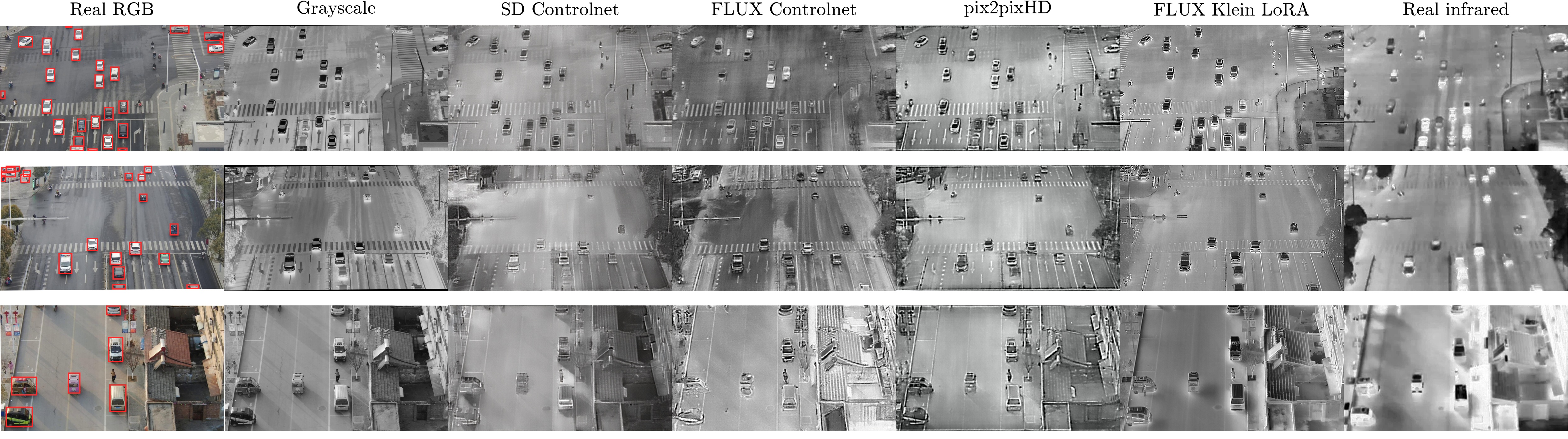}
\caption{Three RGB-to-IR translation example scenes of the VTUAV training set. From left to right, each row shows an RGB image, the translations generated by the grayscale baseline, SD-ControlNet, FLUX-ControlNet, pix2pixHD, and FLUX-LoRA, and the corresponding real infrared image.}
\label{fig:vtuav_inference_examples}
\end{figure}

Figures~\ref{fig:kust4k_inference_examples} and~\ref{fig:vtuav_inference_examples} show example translations on the held-out Kust4K and VTUAV target datasets. Overall, all generative methods preserve the geometric structure of the input image well. Vehicles, road layouts, vegetation, and buildings remain spatially consistent with the RGB input. For Kust4K, the differences between the grayscale baseline and the learned translation methods are particularly visible in the low-light example shown in the second row of Figure~\ref{fig:kust4k_inference_examples}. Because grayscale conversion directly preserves visible-spectrum intensities, the resulting image remains unrealistically dark and some vehicles become difficult to distinguish. The learned RGB-to-IR models generally introduce stronger thermal contrast, although this varies between methods and examples; in the low-light Kust4K example, FLUX-LoRA preserves the visibility of the dark vehicles particularly well. The generated infrared imagery exhibits substantial appearance variation across methods. For example, roads surfaces may appear either relatively hot or relatively cold, while vehicles and surrounding vegetation also vary in thermal contrast.

 
\begin{figure}
\centering
\includegraphics[width=\linewidth]{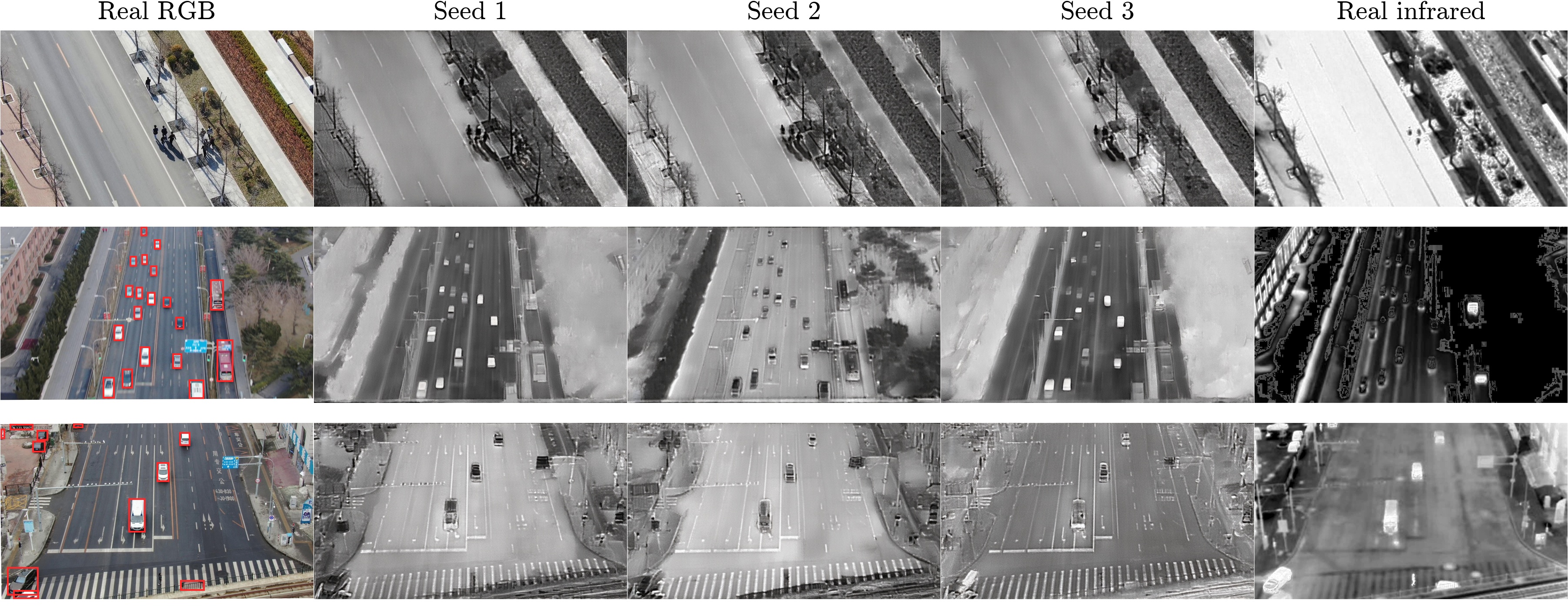}
\caption{Effect of diffusion seed variation for SD-ControlNet on the VTUAV training set. From left to right, each row shows an RGB image, SD-ControlNet outputs generated with different random seeds, and the real infrared target.}
\label{fig:vtuav_sd_seed_examples}
\end{figure}

\begin{figure}
\centering
\includegraphics[width=\linewidth]{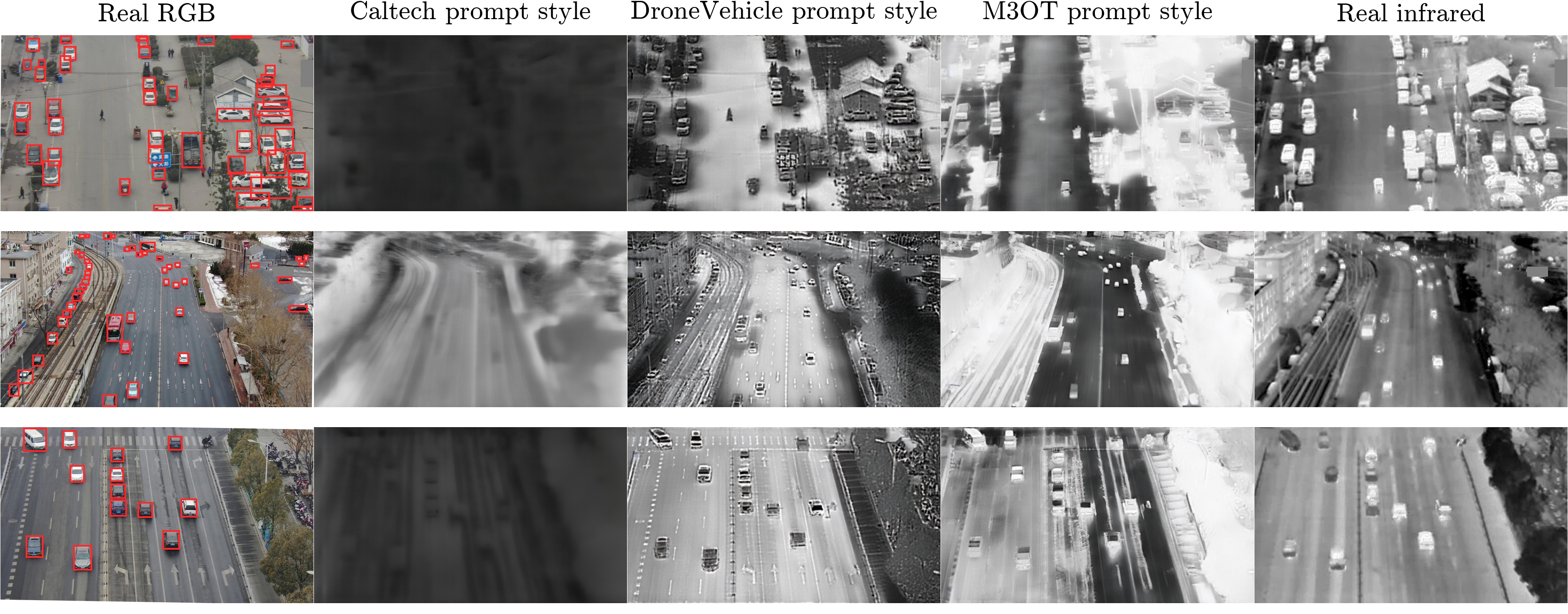}
\caption{Effect of dataset-specific prompt styles for SD-ControlNet on the VTUAV training set. From left to right, each row shows an RGB image, the SD-ControlNet translations generated using prompts corresponding to the three source datasets (DroneVehicle, Caltech, and M3OT), and the real infrared target.}
\label{fig:vtuav_sd_promptstyle_examples}
\end{figure}

Figure~\ref{fig:vtuav_sd_seed_examples} illustrates the effect of varying the diffusion seeds for SD-ControlNet. While object geometry and scene layout remain unchanged, the thermal appearance varies across generations. In particular, the seeds differ in the relative temperature of the road and vehicles: some generations produce warmer backgrounds with comparatively cooler vehicles, while others increase the vehicle-to-background contrast. Figure~\ref{fig:vtuav_sd_promptstyle_examples} shows the effect of dataset-specific prompt styles. The Caltech style produces weak vehicle contrast. The DroneVehicle style resembles the default SD-ControlNet appearance most closely. In contrast, the M3OT style produces darker backgrounds and brighter foreground objects, reflecting characteristics observed in the M3OT dataset and resulting in an appearance that is qualitatively closer to the real VTUAV infrared imagery.

\subsection{Object detection}
Tables~\ref{tab:rfdetr_ood_invert_repeat_results} and~\ref{tab:rfdetr_ood_invert_combi_results} summarize the object detection results. Without any target-domain data, performance is substantially lower on VTUAV ($25.6$ mAP) than on Kust4K ($50.8$ mAP). Adding the real infrared target-domain training data yields large improvements on both datasets, increasing performance to $67.5$ and $52.3$ mAP, respectively. 

Adding target-domain RGB images improves performance by $+6.3$ and $+4.5$ mAP on Kust4K and VTUAV, respectively. Converting these images to grayscale provides a further improvement, resulting in total gains of $+8.5$ and $+8.1$ mAP.

\begin{figure}
    \centering

    \begin{subfigure}[t]{0.32\linewidth}
        \centering
        \includegraphics[width=\linewidth]{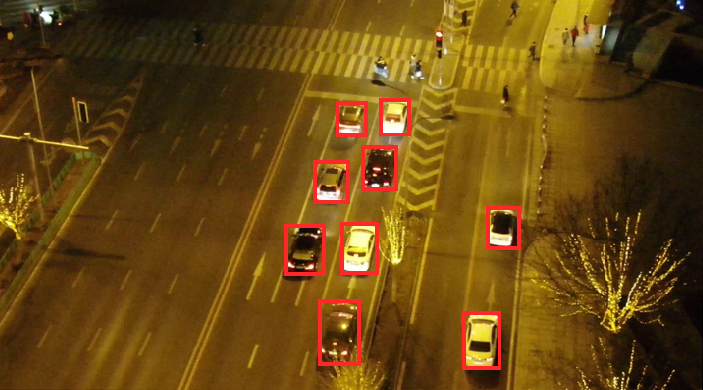}
        \caption{RGB input}
    \end{subfigure}
    \hfill
    \begin{subfigure}[t]{0.32\linewidth}
        \centering
        \includegraphics[width=\linewidth]{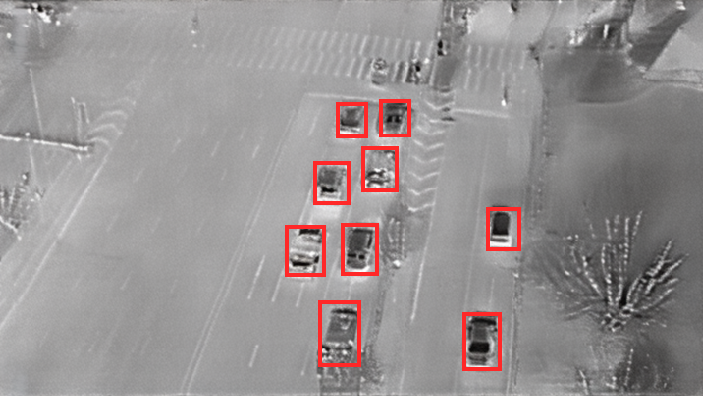}
        \caption{SD-ControlNet}
    \end{subfigure}
    \hfill
    \begin{subfigure}[t]{0.32\linewidth}
        \centering
        \includegraphics[width=\linewidth]{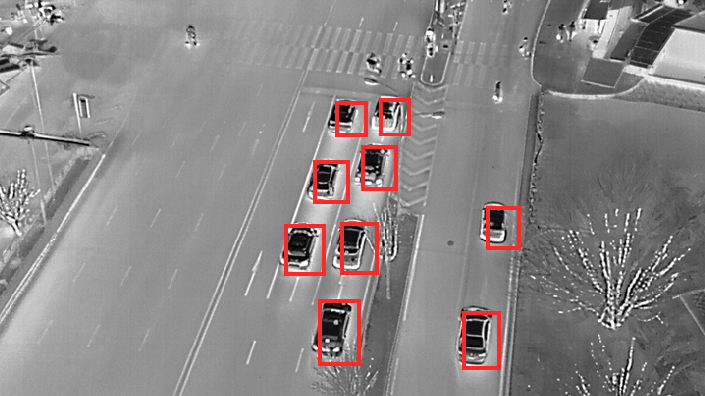}
        \caption{FLUX-LoRA}
    \end{subfigure}

    \caption{Example of spatial inconsistency in the FLUX-LoRA translation. 
    The bounding boxes correspond to the original RGB annotations. 
    SD-ControlNet preserves the vehicle locations, while FLUX-LoRA shifts the vehicles, causing the transferred annotations to become misaligned.}
    \label{fig:klein_spatial_shift}
\end{figure}

Among the learned translation methods, SD-ControlNet performs best overall, improving the source-only baseline by $+9.2$ mAP on Kust4K and $+11.7$ mAP on VTUAV with a single generated variant. Using three seeds provides little additional benefit on Kust4K but increases the VTUAV gain to $+12.8$ mAP. Dataset-specific prompt styles further improve VTUAV to $+15.0$ mAP, while providing no additional benefit on Kust4K. FLUX-ControlNet is less effective on Kust4K but remains competitive on VTUAV, particularly when using three seeds. pix2pixHD provides smaller gains on both datasets, whereas FLUX-LoRA reduces performance below the source-only baseline. As illustrated in Figure~\ref{fig:klein_spatial_shift}, FLUX-LoRA can alter the position of vehicles during translation, causing the transferred RGB annotations to become misaligned.

Performance on all in-domain source test sets remains largely stable across the target-domain augmentation experiments. The source-only baseline achieves $63.3{\pm}0.1$ mAP. Adding real target-domain infrared data improves this performance to $63.8{\pm}0.0$ mAP, while the RGB, grayscale, and synthetic infrared variants lead to only small decreases, with a maximum reduction of $0.4$ mAP.

\begin{table}
\caption{Effect of adding Kust4K and VTUAV target-domain training data to the RF-DETR multi-source baseline.
Values are mean $\pm$ standard deviation over three completed runs, multiplied by 100. Delta values ($\Delta$) are reported relative to the multi-source baseline (\textit{“None”}).}
\label{tab:rfdetr_ood_invert_repeat_results}
\centering
\scriptsize
\setlength{\tabcolsep}{6pt}
\begin{tabular}{@{}lllll@{}}
\toprule
\textbf{Test dataset}        & \multicolumn{2}{c}{Kust4k} & \multicolumn{2}{c}{VTUAV} \\ \midrule
\textbf{Added training dataset} &
  \multicolumn{1}{c}{\textit{$mAP ~ [\pm std]$}} &
  \multicolumn{1}{c}{\textit{$\Delta mAP$}} &
  \multicolumn{1}{c}{\textit{$mAP ~ [\pm std]$}} &
  \multicolumn{1}{c}{\textit{$\Delta mAP$}} \\ \midrule
\textit{None} & $50.8 ~ [\pm 0.1]$ & $0.0$ & $25.6 ~ [\pm 0.3]$ & $0.0$ \\ 
Real RGB & $57.1 ~ [\pm 1.0]$ & $+6.3$ & $30.1 ~ [\pm 0.3]$ & $+4.5$ \\
Grayscale & $59.3 ~ [\pm 0.7]$ & $+8.5$ & $33.7 ~ [\pm 1.4]$ & $+8.1$ \\ 
SD-ControlNet & $60.0 ~ [\pm 0.2]$ & $+9.2$ & $37.3 ~ [\pm 0.8]$ & $+11.7$ \\ 
SD-ControlNet, three seeds & $60.1 ~ [\pm 0.5]$ & $+$9.3 & $38.4 ~ [\pm 0.4]$ & $+12.8$ \\ 
SD-ControlNet, prompt styles & $59.3 ~ [\pm 0.2]$ & $+8.5$ & $40.6 ~ [\pm 0.2]$ & $+$15.0 \\ 
FLUX-ControlNet & $54.8 ~ [\pm 1.8]$ & $+4.0$ & $38.6 ~ [\pm 0.6]$ & $+13.0$ \\ 
FLUX-ControlNet, three seeds & $57.1 ~ [\pm 0.4]$ & $+6.3$ & $39.3 ~ [\pm 0.7]$ & $+13.7$ \\ 
pix2pixHD & $56.3 ~ [\pm 0.2]$ & $+5.5$ & $32.0 ~ [\pm 0.5]$ & $+6.4$ \\ 
FLUX-LoRA & $46.3 ~ [\pm 1.3]$ & $-4.5$ & $21.1 ~ [\pm 2.3]$ & $-4.5$ \\ \midrule 
Real infrared & $67.5 ~ [\pm 0.4]$ & $+16.7$ & $52.3 ~ [\pm 0.3]$ & $+26.7$ 
\\ \bottomrule
\end{tabular}
\end{table}

Table~\ref{tab:rfdetr_ood_invert_combi_results} reports the object detection results when grayscale and SD-ControlNet are combined. Combining grayscale with SD-ControlNet improves over grayscale alone in all settings. On Kust4K, the strongest result is obtained with the simplest SD-ControlNet configuration, increasing performance from 59.3 to 62.4 mAP, while additional seeds or prompt styles do not provide further gains. On VTUAV, all combinations perform similarly at around 40 mAP, suggesting that the main benefit comes from combining the two representations rather than from adding more generative variation.


\begin{table} \caption{Effect of combining grayscale and SD-ControlNet target-domain training data on the RF-DETR multi-source baseline. Values are mean $\pm$ standard deviation over three completed runs, multiplied by 100. Delta values ($\Delta$) are reported relative to the \textit{“Grayscale”} baseline.} \label{tab:rfdetr_ood_invert_combi_results} \centering \scriptsize \setlength{\tabcolsep}{6pt} \begin{tabular}{@{}lllll@{}} \toprule \textbf{Test dataset} & \multicolumn{2}{c}{Kust4K} & \multicolumn{2}{c}{VTUAV} \\ \midrule \textbf{Added training dataset} & \multicolumn{1}{c}{\textit{$mAP ~ [\pm std]$}} & \multicolumn{1}{c}{\textit{$\Delta mAP$}} & \multicolumn{1}{c}{\textit{$mAP ~ [\pm std]$}} & \multicolumn{1}{c}{\textit{$\Delta mAP$}} \\ \midrule 
\textit{Grayscale} & $59.3 ~ [\pm 0.7]$ & $0.0$ & $33.7 ~ [\pm 1.4]$ & $0.0$ \\ 
Grayscale + SD-ControlNet & $62.4 ~ [\pm 0.3]$ & $+3.1$ & $40.2 ~ [\pm 0.1]$ & $+6.5$ \\ 
Grayscale + SD-ControlNet, three seeds & $61.4 ~ [\pm 0.1]$ & $+2.1$ & $39.8 ~ [\pm 0.2]$ & $+6.1$ \\ 
Grayscale + SD-ControlNet, prompt styles & $61.2 ~ [\pm 0.5]$ & $+1.9$ & $41.6 ~ [\pm 1.1]$ & $+7.4$ \\ 
\midrule Real infrared & $67.5 ~ [\pm 0.4]$ & $+8.2$ & $52.3 ~ [\pm 0.3]$ & $+18.6$ \\ \bottomrule \end{tabular} \end{table}

%% file: sections/05_discussion.tex
\section{Discussion}
\label{sec:discussion}
In this work, we investigated the extent to which synthetic infrared imagery, generated from target-domain RGB images using RGB-to-IR translation, can be used as a form of data augmentation in the absence of enough target-domain infrared training images.

\subsection{Benefits of target-domain data under domain shift}
The source-only baseline already highlights the different levels of domain shift between the two target datasets. The detector achieves 50.8 mAP on the close-domain Kust4K, and performance drops to 25.6 mAP on the far-domain VTUAV. This confirms that VTUAV constitutes the more challenging, far-domain, transfer scenario. Adding target-domain RGB images improves performance by +6.3 and +4.5 mAP, respectively, while grayscale conversion increases these gains to +8.5 and +8.1 mAP. This indicates that a substantial part of the benefit comes from exposing the detector to target-domain scene content, including object scales, viewpoints, backgrounds, and annotation distributions. 

The strong performance of grayscale conversion suggests that part of the domain gap is unrelated to infrared appearance. By introducing target-domain geometry and scene content without requiring a learned translation model, grayscale imagery already provides most of the genuinely new information available from the target domain. This observation is consistent with the fact that both the detector and the translation models learn their notion of infrared appearance from the same source-domain infrared datasets (Figure \ref{fig:overview_pipeline}). A translation method may render a vehicle with a more realistic infrared appearance, but similar infrared vehicle appearances have already been seen in the source-domain infrared training data.


Nevertheless, learned RGB-to-IR translation provides an additional benefit. SD-ControlNet with three seeds improves performance by +9.3 mAP on Kust4K and +12.8 mAP on VTUAV, outperforming both the RGB and grayscale baselines. The improvement over grayscale is small on Kust4K (+0.8 mAP), but substantially larger on VTUAV (+4.7 mAP). This suggests that learned translation, and thus more realistic infrared appearance, becomes increasingly important as the domain shift grows. Although using real target-domain infrared remains substantially better (+16.7 and +26.7 mAP), synthetic target-domain infrared recovers a part of the performance lost under domain shift using only target-domain RGB imagery.

The low performance of pix2pixHD may be related to the inherently one-to-many nature of RGB-to-IR translation. Visible appearance does not uniquely determine infrared appearance: from RGB imagery alone, it is generally unknown which objects or object parts are warm and how strongly they contrast with the background. In our paired setup, pix2pixHD learns a predominantly deterministic mapping from RGB to infrared and therefore tends toward a single source-derived translation for a given visible input. 

FLUX-LoRA performs substantially worse than the other translation methods, reducing detection performance to below the source-only baseline. Qualitative inspection suggests that the main problem is spatial inconsistency: the model sometimes changes the position of vehicles instead of only translating their appearance (Figure~\ref{fig:klein_spatial_shift}). Since the original RGB bounding boxes are reused for the generated images, these changes introduce incorrect labels, which is especially problematic for the small objects in aerial imagery. This does not necessarily mean that foundation-model image editing is unsuitable for RGB-to-IR translation. Clouser et al.~\cite{clouser2026fewshot} successfully use a LoRA-adapted foundation model for cross-spectral translation and downstream detection. In our setting, however, stronger preservation of the input geometry appears to be necessary.

\subsection{Role of infrared appearance and generative diversity}
Generating multiple translations with different random seeds provides only a limited benefit. On Kust4K, increasing the number of SD-ControlNet seeds from one to three has a negligible effect (60.0 versus 60.1 mAP), indicating that the close-domain setting already contains sufficient variability. On VTUAV, however, performance slightly improves from 37.3 to 38.4 mAP. This suggests that exposing the detector to multiple plausible infrared appearances might be useful when the domain gap increases, although random seed variation does not provide explicit control over which infrared appearance characteristics are changed.


Variation can be created in a more controlled manner by dataset-specific prompt conditioning, and a larger improvement is obtained in this case. By associating each source dataset with a different prompt during training, the model learns several dataset-associated infrared styles that can be selected at inference time. This also gives less represented source styles a more explicit role at inference, rather than allowing the appearance of the much larger DroneVehicle dataset to dominate the translation. Using the prompt styles increases VTUAV performance to 40.6 mAP, suggesting that explicitly varying the learned infrared style is more useful than relying on random seed variation alone. Qualitatively, the M3OT-associated style produces an infrared appearance that resembles VTUAV more closely, despite substantial differences in viewpoint, object scale, and scene content (Figure~\ref{fig:vtuav_sd_promptstyle_examples}).  This indicates that infrared characteristics learned from one dataset can remain useful even when applied to very different RGB scenes. In contrast, the Caltech-associated style often produces weaker vehicle-to-background contrast, which may reflect the limited similarity between the Caltech training imagery and the highway scenes present in VTUAV. 

The combination experiments suggest that grayscale and SD-translated infrared are to a certain extent complementary methods. Grayscale images provide the target-domain geometry and scene content without introducing translation artifacts, while SD-ControlNet adds an infrared-like appearance learned from the source datasets. This combination is more effective than either representation alone, particularly on VTUAV. However, increasing the number of translated variants through multiple seeds or prompt styles provides little or inconsistent additional improvement once grayscale is included. This suggests that most of the benefit comes from combining grayscale and translated infrared representations, rather than simply adding more generated samples.

\subsection{Limitations and future work}
A fundamental limitation of RGB-to-IR translation is that thermal appearance cannot be uniquely inferred from RGB imagery. Effects such as engine heat, object temperature, and environmental thermal history are not directly visible, so the translator can only reproduce thermal patterns learned from the source data. This is further limited by the relatively small and imbalanced set of source infrared datasets, which is dominated by DroneVehicle. A larger and more balanced collection could expose the model to a wider range of sensors, day/night conditions, object thermal states, and other infrared appearances. An alternative route is to add a limited amount of in-domain real data to the synthetic data, as was shown to be an effective strategy for vehicle detection in RGB imagery \cite{Heslinga2024combining, Heslinga2024Simulation, snel2025data}. 


Future work should therefore focus on representing thermal variation more explicitly. The prompt-style experiment suggests that controlled variation can be more useful than relying on random seed variation alone, and this could be extended beyond dataset-level styles toward specific thermal characteristics such as sensor type, time of day, road temperature, or vehicle operating state. At the same time, grayscale was deliberately kept as a simple baseline in this study. For a fairer comparison with increasingly sophisticated generative methods, future work should also investigate stronger non-generative alternatives, such as semantic-aware grayscale augmentation~\cite{d2025saga}.

Finally, the evaluation is limited to two unseen UAV target domains. Additional datasets covering different sensors, environments, flight configurations, and thermal conditions would be needed to determine how broadly the findings generalize. Real target-domain infrared data still substantially outperforms all synthetic alternatives. RGB-to-IR translation should therefore be viewed as an intermediate strategy: when target-domain RGB imagery and annotations are available but corresponding infrared training data are scarce or unavailable, synthetic translation can recover a substantial part of the lost cross-domain detection performance.